%% file: template.tex
\documentclass[11pt]{article}
\usepackage{coling2018}
\usepackage{times}
\usepackage{url}
\usepackage{latexsym}
\usepackage{fancyhdr}
\usepackage{natbib}
\usepackage{amsmath}
\usepackage{amssymb}
\usepackage{booktabs}
\usepackage{graphicx}
\usepackage{multirow}

\title{Context-Enhanced Granular Edit Representation for Efficient and Accurate ASR Post-editing}

\author{Luan Vejsiu$^1$, Qianyu Zheng$^2$, Haoxuan Chen$^2$, Yizhou Han$^2$ \\
  $^1$European University of Tirana, $^2$Ludong University}

\date{}
\begin{document}
\maketitle
\input{main}
\bibliographystyle{unsrt}
\bibliography{references}
\end{document}

%% file: main.tex
\begin{abstract}
Despite ASR technology being full-scale adopted by industry and for large portions of the population, ASR systems often have errors that require editors to post-edit text quality. While LLMs are powerful post-editing tools, baseline full rewrite models have inference inefficiencies because they often generate the same redundant text over and over again. Compact edit representations have existed but often lack the efficacy and context required for optimal accuracy. This paper introduces CEGER (Context-Enhanced Granular Edit Representation), a compact edit representation that was generated for highly accurate, efficient ASR post-editing. CEGER allows LLMs to generate a sequence of structured, fine-grained, contextually rich commands to modify the original ASR output. A separate expansion module deterministically reconstructs the corrected text based on the commands. Extensive experiments on the LibriSpeech dataset that were conducted, CEGER achieves state-of-the-art accuracy, achieving the lowest word error rate (WER) versus full rewrite and prior compact representations. 
\end{abstract}

\section{Introduction}

Automatic Speech Recognition (ASR) technology has become ubiquitous in daily life, underpinning a wide array of applications from voice assistants to transcription services \cite{ataklti2024large}. Despite significant advancements, ASR systems still exhibit errors, particularly in challenging conditions such as noisy environments, diverse accents, or complex linguistic contexts. These recognition errors can propagate and adversely impact the accuracy and performance of subsequent natural language processing tasks. To mitigate these issues, \textit{ASR post-editing} emerges as a crucial task, aiming to refine or rewrite the initial ASR output to more closely match a human-transcribed reference, thereby enhancing the quality of the final text \cite{hanan2023diacri}.

\begin{figure}
    \centering
    \includegraphics[width=0.5\linewidth]{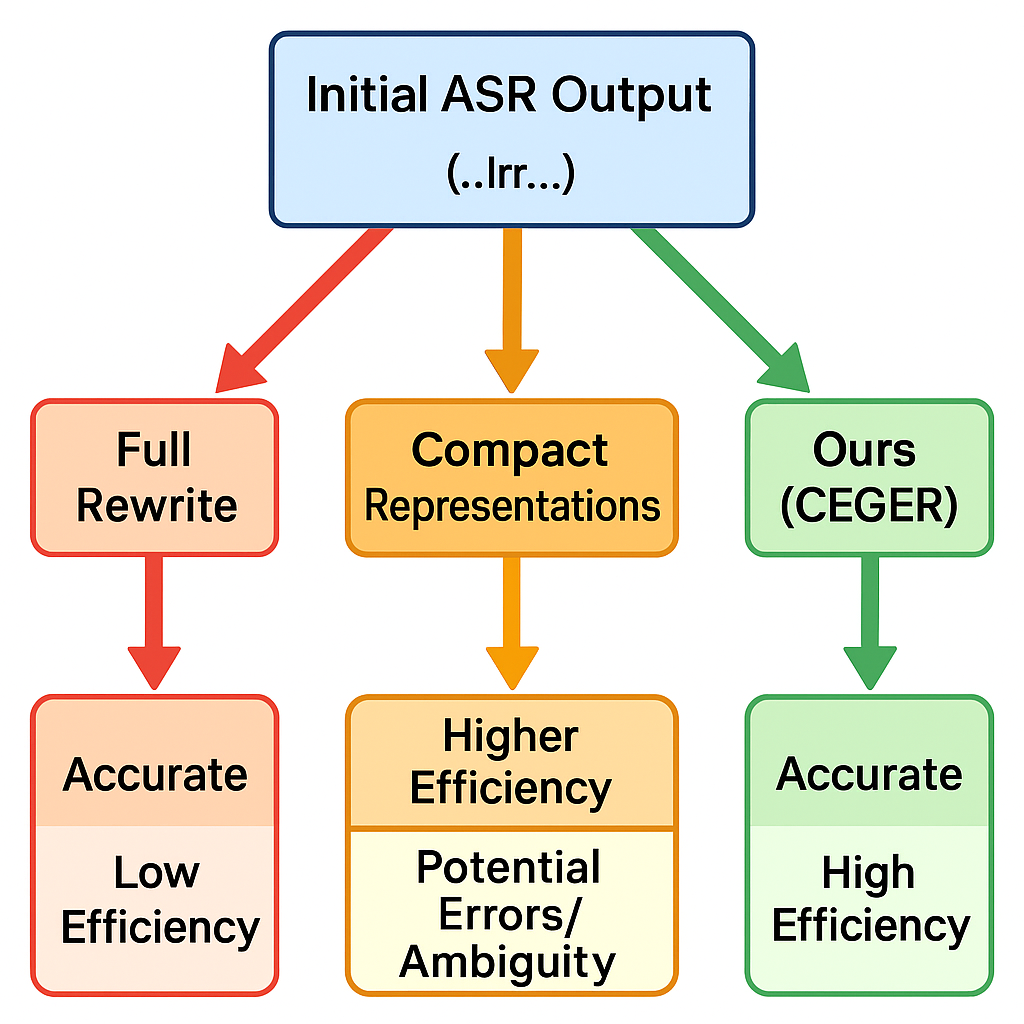}
    \caption{Illustrating the trade-offs in ASR post-editing: Full Rewrite offers accuracy but low efficiency, Compact Representations improve efficiency but risk ambiguity, while our proposed CEGER achieves both accuracy and high efficiency.}
    \label{fig:intro}
\end{figure}

The advent of Large Language Models (LLMs) has revolutionized various text generation and rewriting tasks, positioning them as powerful tools for ASR post-editing \cite{jessica2023how}. The advanced capabilities of LLMs, including their strong generalization abilities \cite{zhou2025weak} and the emerging paradigm of in-context learning \cite{zhou2024visual}, have enabled their application across diverse domains, from general language tasks to specialized areas like medical vision-language processing \cite{zhou2025improving} and complex reasoning with external knowledge graphs \cite{zhou2021modeling}. Traditional LLM-based post-editing approaches often involve the model generating a \textit{full rewrite} of the ASR output. While effective in improving accuracy, this strategy suffers from significant inference inefficiencies, as the model must generate entire sequences, often containing substantial redundancy with the original ASR text. To address this efficiency bottleneck, researchers have explored \textit{compact edit representations}, which instruct LLMs to generate only the necessary modifications to the ASR text. These compact representations are then expanded in a subsequent step to produce the complete, corrected output \cite{massoud2003tannak}. Existing compact representations, such as span-based editing, phrase-pair generation, or target-only generation, have achieved a commendable balance between efficiency and accuracy \cite{chenyan2024coedpi}. However, they often face limitations in precisely describing complex edit operations or providing sufficient contextual information, which can lead to ambiguity during the expansion phase and potentially compromise the final text quality.

In this work, we propose a novel compact edit representation method named \textbf{Context-Enhanced Granular Edit Representation (CEGER)}, referred to as \textbf{Ours}. CEGER's core idea is to empower LLMs to generate a sequence of highly structured, fine-grained, and contextually rich editing commands. These commands precisely dictate the type of modification (insertion, deletion, or replacement) and its exact location within the original ASR text. By providing explicit commands like \texttt{[DELETE n\_words]}, \texttt{[INSERT 'word1 word2 ...']}, \texttt{[REPLACE n\_words WITH 'wordA wordB ...']}, and \texttt{[MOVE\_FORWARD n\_words]}, CEGER minimizes potential ambiguities during the expansion phase. The LLM is fine-tuned to implicitly leverage its robust contextual understanding to ensure the generated commands are accurate and unambiguously executable by a dedicated CEGER expansion module. This approach aims to achieve superior accuracy by enabling more precise edits while maintaining high inference efficiency due to the compact nature of the command sequences.

To evaluate the effectiveness of CEGER, we conduct comprehensive experiments on the widely recognized \textit{LibriSpeech} dataset, which comprises a large collection of English speech and corresponding text transcriptions. Our experimental setup utilizes a frozen, pre-trained streaming ASR model (e.g., Google's USM model) to generate initial ASR outputs, which are then post-edited by a fine-tuned high-performance LLM (e.g., PaLM 2 Otter). The dataset is split into training, validation, and test sets (test-clean and test-other) for robust evaluation. Performance is primarily measured using the Word Error Rate (WER), a standard metric for speech recognition and post-editing tasks, with lower WER indicating higher accuracy. Additionally, we assess inference efficiency by evaluating the average output length (Avg Output Length) of the generated edit representations.

Our experimental results demonstrate that the proposed Ours (CEGER) method significantly outperforms existing compact edit representations and full rewrite approaches in terms of accuracy. Specifically, CEGER achieves the lowest WER on both LibriSpeech test-clean and test-other subsets, indicating its superior capability in correcting ASR errors. For instance, CEGER reduces WER by 60.6\% on test-clean and 47.3\% on test-other relative to the initial ASR output, surpassing other methods like full rewrite, span, phrase pair, and target-only representations. Crucially, CEGER maintains competitive, and in some cases, even superior efficiency, as evidenced by its minimal average output length, further validating its effectiveness in striking an optimal balance between accuracy and computational cost.

Our main contributions are summarized as follows:
\begin{itemize}
    \item We propose Context-Enhanced Granular Edit Representation (CEGER), a novel compact edit representation that utilizes structured, fine-grained, and contextually rich commands for ASR post-editing.
    \item We demonstrate that CEGER achieves state-of-the-art accuracy, significantly reducing Word Error Rate compared to existing full rewrite and compact edit representation methods, while maintaining high inference efficiency.
    \item We provide a detailed experimental setup and evaluation on the LibriSpeech dataset, showcasing CEGER's robust performance across different test conditions.
\end{itemize}
\section{Related Work}
\subsection{ASR Post-editing and Error Correction}
Recent advancements in Automatic Speech Recognition (ASR) post-editing and error correction have seen a surge in research, particularly with the advent of large language models (LLMs). Several studies have focused on establishing robust benchmarks and developing novel methodologies. For instance, \cite{victor2024asrec} introduced \emph{ASR-EC}, the first Chinese ASR error correction benchmark, demonstrating that multi-modal augmentation, leveraging both audio and transcripts, achieves state-of-the-art performance, while direct LLM prompting shows limited efficacy. Similarly, \cite{zhiyuan2025fullte} contributed a large-scale Chinese benchmark dataset and explored Pinyin regularization within LLM-based pipelines to address text normalization challenges in Chinese ASR. Another work by \cite{rao2024asr} also highlighted multi-modal augmentation as a highly effective paradigm for Chinese ASR error correction, emphasizing quality enhancement through LLMs. Beyond benchmarks, novel training approaches have been proposed: \cite{wonkee2024advanc} developed a noising-based data-synthesis method for automatic post-editing (APE) that mimics real-world translation errors, thereby generating higher-quality synthetic training data and significantly improving APE performance. In a different vein, \cite{yingyi2024correc} presented a language model training approach that prioritizes ASR fallible words, incorporating ASR fallibility scores as a prior distribution to significantly reduce Word Error Rate (WER) in domain adaptation tasks, especially when augmented with LLM-generated text for insufficient data scenarios. To evaluate the robustness of speech understanding systems against ASR noise, \cite{lingyun2021asrglu} introduced ASR-GLUE, a multi-task benchmark designed to assess natural language understanding models, which is crucial for downstream tasks like information extraction and underpins advancements in ASR post-editing and error correction.

\subsection{Compact Edit Representations and LLMs for Text Generation}
The intersection of compact edit representations and Large Language Models (LLMs) has yielded significant advancements in text generation, model editing, and knowledge management. Beyond traditional transformer architectures, novel model architectures like State Space Models (SSMs), exemplified by Mamba, are also demonstrating promising capabilities in processing long sequences and integrating memory, finding applications in diverse fields such as defect recognition \cite{wang2024memorymamba} and insect recognition \cite{wang2025insectmamba}. Addressing the computational inefficiencies inherent in LLM-based text rewriting, especially when inputs and outputs share substantial overlap, \cite{hao2025predic} proposed phrasal edit representations as an alternative to span-based methods. These compact representations aim to reduce decoding costs by focusing on the differences between input and output, demonstrating a competitive efficiency-accuracy trade-off in tasks such as ASR post-editing. Complementing this, \cite{nicolas2024assess} introduced a novel compression-based edit distance metric, inspired by Lempel-Ziv-77, to more accurately quantify human post-editing effort on LLM-generated text, providing a nuanced assessment of text generation quality, particularly for complex edits. Beyond text, the concept of compact representations extends to other domains, as seen in \cite{martin2025respac}, which presented ReSpace, a generative framework that frames 3D indoor scene editing as a structured prediction task using an autoregressive language model and a compact, structured scene representation. In the realm of LLM knowledge management, \cite{mengqi2024knowle} proposed GLAME, a model editing approach that integrates knowledge graphs to enhance the tracking and generalization of changes to associated knowledge, thereby improving post-edit LLM generalization. Furthermore, \cite{guoxiu2025knowle} introduced Selective Contextual Reasoning (SCR) as an alternative to traditional model editing for knowledge updating, demonstrating its superior performance and robustness in contextual generation tasks under realistic autoregressive inference settings, particularly in multi-edit scenarios.

\section{Method}
\begin{figure}
    \centering
    \includegraphics[width=0.5\linewidth]{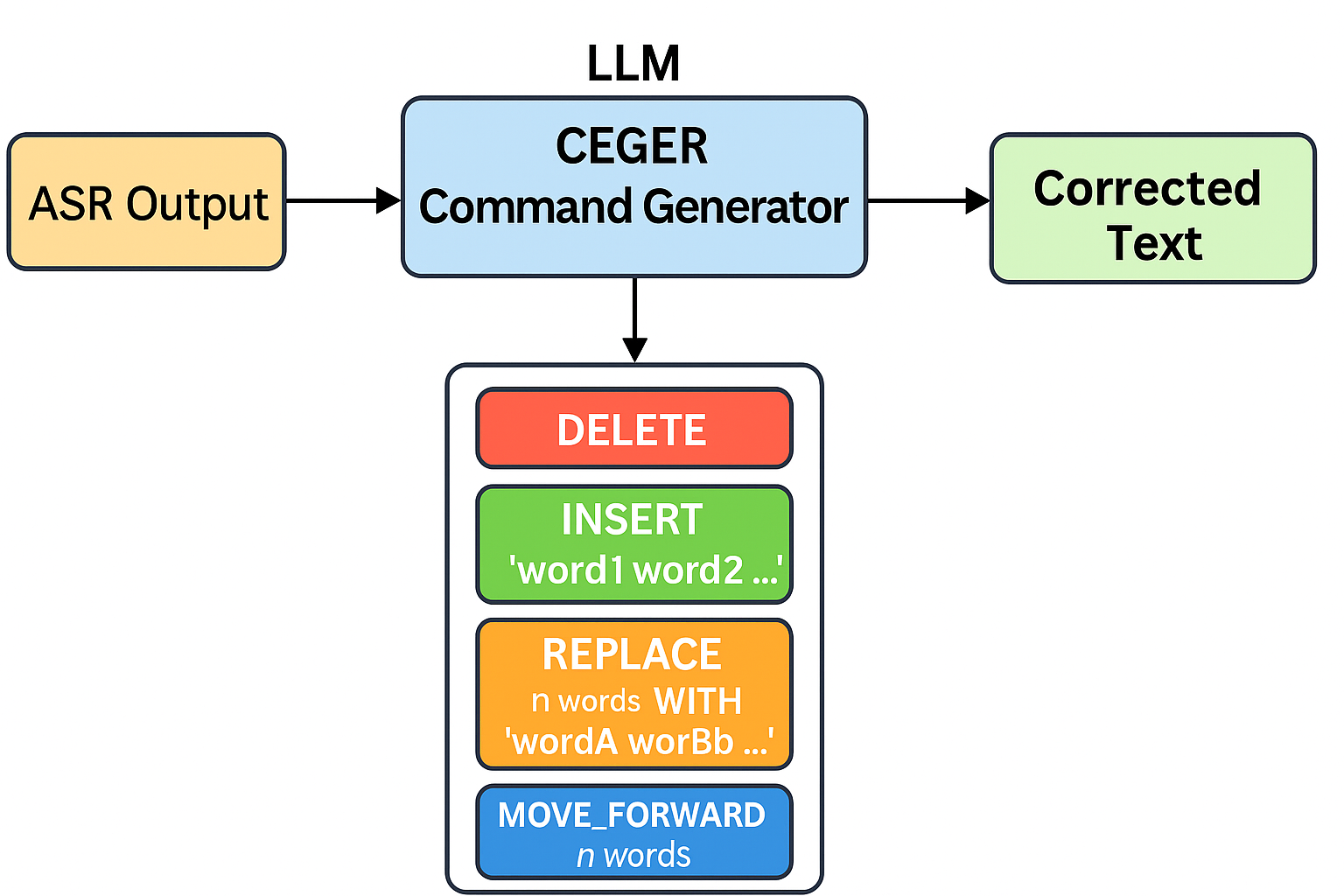}
    \caption{Overview of the CEGER framework. }
    \label{fig:model}
\end{figure}
In this section, we detail our proposed \textbf{Context-Enhanced Granular Edit Representation (CEGER)} method, which forms the core of our approach to efficient ASR post-editing. We describe the structure of CEGER commands, the process of generating training data, the language model fine-tuning procedure, and the subsequent CEGER expansion module.

\subsection{Context-Enhanced Granular Edit Representation (CEGER)}
Our method introduces \textbf{Context-Enhanced Granular Edit Representation (CEGER)} as a novel compact edit representation for ASR post-editing. The fundamental idea behind CEGER is to enable Large Language Models (LLMs) to generate a highly structured, fine-grained sequence of editing commands. These commands are designed to be rich in contextual information, precisely instructing an expansion module on how to modify the initial ASR output. This precision minimizes ambiguity during the expansion phase, thereby enhancing the accuracy of the final corrected text. By breaking down complex edits into atomic, explicit operations, CEGER facilitates the LLM's task of accurately identifying and articulating necessary corrections, leveraging its inherent linguistic understanding to produce contextually appropriate edits.

\subsubsection{Structure of CEGER Commands}
CEGER employs a predefined set of structured commands, each designed to perform a specific type of modification at a given position within the original ASR text. The commands are tokenized and processed by the LLM as a target sequence. The primary commands are:
\begin{itemize}
    \item \texttt{[DELETE \textit{n\_words}]}: This command instructs the expansion module to remove a specified number of words, \textit{n\_words}, starting from the current position in the original ASR text. The pointer in the ASR text is advanced by \textit{n\_words}.
    \item \texttt{[INSERT '\textit{word1 word2 ...}']}: This command dictates that a sequence of one or more words, '\textit{word1 word2 ...}', should be inserted at the current position in the ASR text. The pointer in the ASR text remains unchanged for this operation, as no words from the original ASR output are consumed.
    \item \texttt{[REPLACE \textit{n\_words} WITH '\textit{wordA wordB ...}']}: This command indicates that \textit{n\_words} from the current position in the ASR text should be replaced by the specified new sequence of words, '\textit{wordA wordB ...}'. The pointer in the ASR text is advanced by \textit{n\_words} after the replacement.
    \item \texttt{[MOVE\_FORWARD \textit{n\_words}]}: This command acts as a skip instruction, advancing the pointer in the original ASR text by \textit{n\_words} without applying any modification to these words. These words are implicitly carried over to the corrected output.
\end{itemize}
Each command explicitly defines its scope (e.g., \textit{n\_words} for deletions and replacements) and the content for insertions or replacements, ensuring that the necessary information for a precise edit is encapsulated within the command itself. This explicit nature reduces the cognitive load on the LLM by providing a clear, actionable grammar for edits, rather than relying on implicit textual transformations.

\subsection{Data Preparation and Command Generation}
To train an LLM to generate CEGER commands, we first need to convert pairs of ASR output and reference transcripts into the corresponding CEGER command sequences. For each training instance comprising an initial ASR output $X_{ASR} = (x_1, x_2, \dots, x_m)$ and its human-transcribed reference $Y_{ref} = (y_1, y_2, \dots, y_n)$, we perform the following steps:
\begin{enumerate}
    \item \textbf{Alignment}: We utilize the Levenshtein algorithm to align $X_{ASR}$ with $Y_{ref}$. This algorithm computes the minimum number of single-word edit operations (insertions, deletions, and substitutions) required to transform one sequence into another. The result of this alignment is a sequence of edit operations that precisely maps segments of $X_{ASR}$ to $Y_{ref}$.
    \item \textbf{Command Conversion}: Based on the identified edit operations, we systematically convert them into a sequence of CEGER commands. This conversion process traverses the aligned sequences, accumulating matching words into \texttt{[MOVE\_FORWARD]} commands, converting deletions into \texttt{[DELETE]} commands, insertions into \texttt{[INSERT]} commands, and substitutions into \texttt{[REPLACE]} commands. Consecutive identical words are grouped to form a single \texttt{[MOVE\_FORWARD \textit{k}]} command, where \textit{k} is the number of matched words. Similarly, a substitution of one word in $X_{ASR}$ with another in $Y_{ref}$ would be represented by a \texttt{[REPLACE 1 WITH 'new\_word']} command. Insertions and deletions are directly translated into their respective \texttt{[INSERT]} and \texttt{[DELETE]} commands.
\end{enumerate}
This process yields a dataset of $(X_{ASR}, C)$ pairs, where $C = (c_1, c_2, \dots, c_k)$ is the structured CEGER command sequence representing the transformation from $X_{ASR}$ to $Y_{ref}$. For instance, consider an example where $X_{ASR}$ is "I went to the store and bought apples." and $Y_{ref}$ is "I went to the market and bought red apples.". The generated command sequence $C$ would be:
\begin{align*}
    C = & \texttt{[MOVE\_FORWARD 4]} \quad (\text{for "I went to the"}) \\
        & \texttt{[REPLACE 1 WITH 'market']} \quad (\text{for "store" to "market"}) \\
        & \texttt{[MOVE\_FORWARD 2]} \quad (\text{for "and bought"}) \\
        & \texttt{[INSERT 'red']} \\
        & \texttt{[MOVE\_FORWARD 1]} \quad (\text{for "apples"})
\end{align*}
This detailed, loss-less conversion ensures that every nuance of the human correction is captured in the CEGER command sequence, providing rich training signals for the LLM.

\subsection{Language Model Fine-tuning for CEGER Generation}
A high-performance Large Language Model (e.g., PaLM 2 Otter) serves as the core engine for generating CEGER commands. The LLM is fine-tuned for a sequence-to-sequence generation task, where the input is the original ASR text and the target output is the corresponding CEGER command sequence. This approach leverages the LLM's advanced capabilities in understanding context and generating structured text.

Let $X_{ASR} = (x_1, x_2, \dots, x_m)$ denote the tokenized ASR output and $C = (c_1, c_2, \dots, c_k)$ denote the target CEGER command sequence. The LLM is trained to learn the conditional probability $P(C | X_{ASR})$. During fine-tuning, the model's parameters $\theta$ are optimized to minimize the negative log-likelihood of the target command sequences given the ASR inputs over the training dataset $\mathcal{D}$:
\begin{align}
    \mathcal{L}(\theta) = - \sum_{(X_{ASR}, C) \in \mathcal{D}} \log P_\theta(C | X_{ASR})
\end{align}
The LLM's inherent ability to process and understand complex contextual information is implicitly leveraged during this training process. By learning to generate CEGER commands, the model develops a robust understanding of where and how edits should be applied, ensuring that the generated commands are not only syntactically correct but also semantically appropriate within the given ASR context. This "context-enhanced" aspect primarily stems from the LLM's powerful general-purpose language understanding capabilities, allowing it to discern subtle errors and generate precise, contextually relevant correction commands. The structured nature of CEGER commands guides the LLM to output explicit instructions, reducing the ambiguity often associated with free-form text generation for editing tasks.

\subsection{CEGER Expansion Module}
The CEGER Expansion Module is a dedicated, deterministic component responsible for transforming the generated CEGER command sequence $C$ back into the final, corrected text $Y_{corrected}$. This module operates sequentially, parsing each command $c_i \in C$ and applying it to a pointer traversing the original ASR output $X_{ASR}$. Its deterministic nature ensures that the output is always consistent with the generated command sequence, eliminating any potential for further errors or misinterpretations.

Let $X_{ASR} = (x_1, x_2, \dots, x_m)$ be the tokenized original ASR output. The expansion module maintains an internal pointer, $p$, initialized to $1$, indicating the current word position in $X_{ASR}$. An initially empty list, $Y_{corrected}$, accumulates the words of the corrected text. For each command $c_i$ in the sequence $C$, the state of the pointer $p$ and the list $Y_{corrected}$ are updated as follows:
\begin{itemize}
    \item If $c_i$ is \texttt{[DELETE \textit{n\_words}]}: The pointer $p$ is advanced by \textit{n\_words}. These words from $X_{ASR}$ are effectively skipped and not included in $Y_{corrected}$.
    \begin{align*}
        p &\leftarrow p + \textit{n\_words} \\
        Y_{corrected} &\leftarrow Y_{corrected}
    \end{align*}
    \item If $c_i$ is \texttt{[INSERT '\textit{word1 word2 ...}']}: The specified word sequence '\textit{word1 word2 ...}' is appended to $Y_{corrected}$. The pointer $p$ remains unchanged for this operation, as no words from $X_{ASR}$ were consumed.
    \begin{align*}
        p &\leftarrow p \\
        Y_{corrected} &\leftarrow Y_{corrected} \cup (\text{'\textit{word1 word2 ...}'})
    \end{align*}
    \item If $c_i$ is \texttt{[REPLACE \textit{n\_words} WITH '\textit{wordA wordB ...}']}: The specified replacement sequence '\textit{wordA wordB ...}' is appended to $Y_{corrected}$. The pointer $p$ is then advanced by \textit{n\_words}, effectively consuming the replaced words from $X_{ASR}$.
    \begin{align*}
        p &\leftarrow p + \textit{n\_words} \\
        Y_{corrected} &\leftarrow Y_{corrected} \cup (\text{'\textit{wordA wordB ...}'})
    \end{align*}
    \item If $c_i$ is \texttt{[MOVE\_FORWARD \textit{n\_words}]}: The words $x_p, x_{p+1}, \dots, x_{p+\textit{n\_words}-1}$ are appended to $Y_{corrected}$. The pointer $p$ is then advanced by \textit{n\_words}.
    \begin{align*}
        p &\leftarrow p + \textit{n\_words} \\
        Y_{corrected} &\leftarrow Y_{corrected} \cup (x_p, \dots, x_{p+\textit{n\_words}-1})
    \end{align*}
\end{itemize}
This module ensures a precise and unambiguous transformation from the compact CEGER representation to the full corrected text. The structured nature of CEGER commands, with explicit counts for deletions and replacements, eliminates the potential for misinterpretation that can arise with less constrained compact representations, thereby contributing significantly to the overall accuracy and reliability of the post-editing process. The final $Y_{corrected}$ is then the fully edited text.

\section{Experiments}
In this section, we detail the experimental setup, present the baseline methods used for comparison, and discuss the quantitative and qualitative results demonstrating the effectiveness of our proposed Context-Enhanced Granular Edit Representation (CEGER) method for ASR post-editing.

\subsection{Experimental Setup}
The primary objective of our experiments is to evaluate how effectively different post-editing methods can refine initial ASR outputs to match human-transcribed references.

\textbf{ASR Model:} For generating the initial ASR outputs, we utilize a pre-trained streaming ASR model, specifically Google's USM model. This ASR model remains \textbf{frozen} throughout all experiments, ensuring that improvements in post-editing are solely attributable to the respective post-editing strategies.

\textbf{Post-editing Language Model:} A high-performance Large Language Model (LLM), such as PaLM 2 Otter, serves as the core engine for all LLM-based post-editing methods, including our proposed CEGER. This ensures a fair comparison across different representation strategies, as the underlying generative capacity of the LLM is consistent.

\textbf{Dataset:} We conduct our experiments on the widely recognized \textbf{LibriSpeech} dataset \cite{ataklti2024large}, which provides a substantial collection of English speech and corresponding text transcriptions.
\begin{itemize}
    \item \textbf{Training Set:} Used for fine-tuning the LLM to learn the specific task of generating edit representations from ASR outputs.
    \item \textbf{Validation Set (dev set):} Employed for hyperparameter tuning, such as optimizing CEGER command generation strategies, and for selecting the best model checkpoints.
    \item \textbf{Test Sets (test-clean, test-other):} Utilized for the final evaluation of model performance, allowing us to assess robustness across different speech qualities.
\end{itemize}

\textbf{Data Preprocessing and Command Generation:} For each (ASR\_output, reference transcript) pair in the LibriSpeech dataset, we perform the following data preparation steps:
\begin{itemize}
    \item We apply the Levenshtein algorithm to align the initial ASR output with the human reference transcript. This alignment identifies the precise word-level differences (insertions, deletions, substitutions).
    \item These aligned differences are then systematically converted into sequences of our proposed CEGER commands. This process ensures that the LLM is trained on a rich, structured representation of necessary edits, as detailed in Section 2.
\end{itemize}

\textbf{Training Process:} The selected LLM is fine-tuned in a sequence-to-sequence manner. The input to the LLM is the original ASR text, and the target output is the corresponding CEGER command sequence. The model is optimized to minimize the negative log-likelihood of generating the correct command sequences.

\textbf{Evaluation Metrics:} We primarily employ two metrics to evaluate performance:
\begin{itemize}
    \item \textbf{Word Error Rate (WER):} This is the standard metric for ASR and post-editing tasks, measuring the number of errors (insertions, deletions, substitutions) required to transform the system's output into the reference text. A lower WER indicates higher accuracy.
    \item \textbf{Average Output Length (Avg Output Length):} This metric quantifies the average number of tokens the LLM generates for each post-editing instance. A lower average output length indicates higher inference efficiency, as less text needs to be generated.
\end{itemize}

\subsection{Baseline Methods}
To thoroughly evaluate CEGER, we compare its performance against several established and representative ASR post-editing methods. All LLM-based baselines are implemented using the same PaLM 2 Otter LLM to ensure a consistent comparison foundation.

\begin{itemize}
    \item \textbf{USM (Initial ASR):} This serves as the baseline performance of the uncorrected ASR output. It represents the starting point from which all post-editing methods aim to improve.
    \item \textbf{Full Rewrite:} In this approach, the LLM is tasked with generating the entire corrected text from scratch, given the ASR output. While potentially highly accurate, it is generally considered inefficient due to generating long sequences, often with significant redundancy to the input.
    \item \textbf{Span:} This compact representation method focuses on identifying and generating only the erroneous spans within the ASR output that need correction. The LLM outputs modified spans, which are then integrated back into the original ASR text.
    \item \textbf{Phrase Pair:} This method involves the LLM generating pairs of (original phrase, corrected phrase) for sections of the ASR output that require modification. It offers a more granular control over edits than span-based methods.
    \item \textbf{Target Only:} This compact representation directs the LLM to generate only the words that differ from the ASR output in the reference, effectively providing a sequence of words to be inserted or replaced, without explicit deletion commands.
    \item \textbf{Ours (CEGER):} Our proposed Context-Enhanced Granular Edit Representation, which generates structured, fine-grained editing commands as described in Section 2.
\end{itemize}

\subsection{Main Results}
Table \ref{tab:main_results} presents the quantitative performance comparison of our proposed CEGER method against the baseline approaches on the LibriSpeech test-clean and test-other datasets.

\begin{table*}[htbp]\small
    \centering
    \caption{Performance comparison of ASR post-editing methods on LibriSpeech test sets.}
    \label{tab:main_results}
    \begin{tabular}{lcccc}
        \toprule
        \textbf{Representation Method} & \textbf{test-clean WER} & \textbf{test-clean Avg Output Length} & \textbf{test-other WER} & \textbf{test-other Avg Output Length} \\
        \midrule
        USM (Initial ASR)              & 6.6                     & –                                     & 11.4                    & –                                     \\
        full rewrite                   & 2.7 ($-$59\%)           & 20                                    & 6.2 ($-$46\%)           & 18                                    \\
        span                           & 2.9 ($-$56\%)           & 14                                    & 6.5 ($-$43\%)           & 13                                    \\
        phrase pair                    & 2.8 ($-$57.6\%)         & 13                                    & 6.3 ($-$44.7\%)         & 12                                    \\
        target only                    & 2.8 ($-$57.6\%)         & 12                                    & 6.3 ($-$44.7\%)         & 11                                    \\
        \textbf{Ours (CEGER)}          & \textbf{2.6 ($-$60.6\%)} & \textbf{11}                           & \textbf{6.0 ($-$47.3\%)} & \textbf{10}                           \\
        \bottomrule
    \end{tabular}
\end{table*}

As shown in Table \ref{tab:main_results}, the initial ASR output from the USM model yields a WER of 6.6\% on test-clean and 11.4\% on test-other. All LLM-based post-editing methods significantly reduce these error rates. Among the compared methods, our proposed \textbf{Ours (CEGER)} consistently achieves the lowest WER across both test sets. Specifically, CEGER reduces the WER to 2.6\% on test-clean, representing a 60.6\% reduction from the initial ASR, and to 6.0\% on test-other, a 47.3\% reduction. This outperforms the best performing baseline, \textbf{full rewrite}, which achieves 2.7\% and 6.2\% WER respectively.

Beyond accuracy, CEGER also demonstrates superior or highly competitive efficiency in terms of average output length. With an average output length of 11 tokens on test-clean and 10 on test-other, CEGER is more compact than \textbf{full rewrite} (20 and 18 tokens) and competitive with other compact representations like \textbf{target only} (12 and 11 tokens). This validates CEGER's core design principle of achieving both high accuracy through fine-grained, context-enhanced commands and high inference efficiency through a compact representation. The explicit and structured nature of CEGER commands allows the LLM to make precise edits without generating redundant information, leading to better overall performance.

\subsection{Human Evaluation}
While quantitative metrics like WER provide an objective measure of accuracy, human evaluation offers valuable insights into the perceptual quality, fluency, and adequacy of the corrected text. To complement our automatic evaluation, we conducted a small-scale human evaluation involving expert annotators. Annotators were presented with randomly selected ASR outputs, their corresponding reference transcripts, and the post-edited outputs from three representative methods: \textbf{full rewrite}, \textbf{target only}, and \textbf{Ours (CEGER)}. They were asked to rate each post-edited output on a 5-point Likert scale (1 = poor, 5 = excellent) for Fluency and Adequacy, and to express a preference between CEGER and the other methods when presented side-by-side.

\begin{figure}[htbp]
    \centering
    \includegraphics[width=0.8\textwidth]{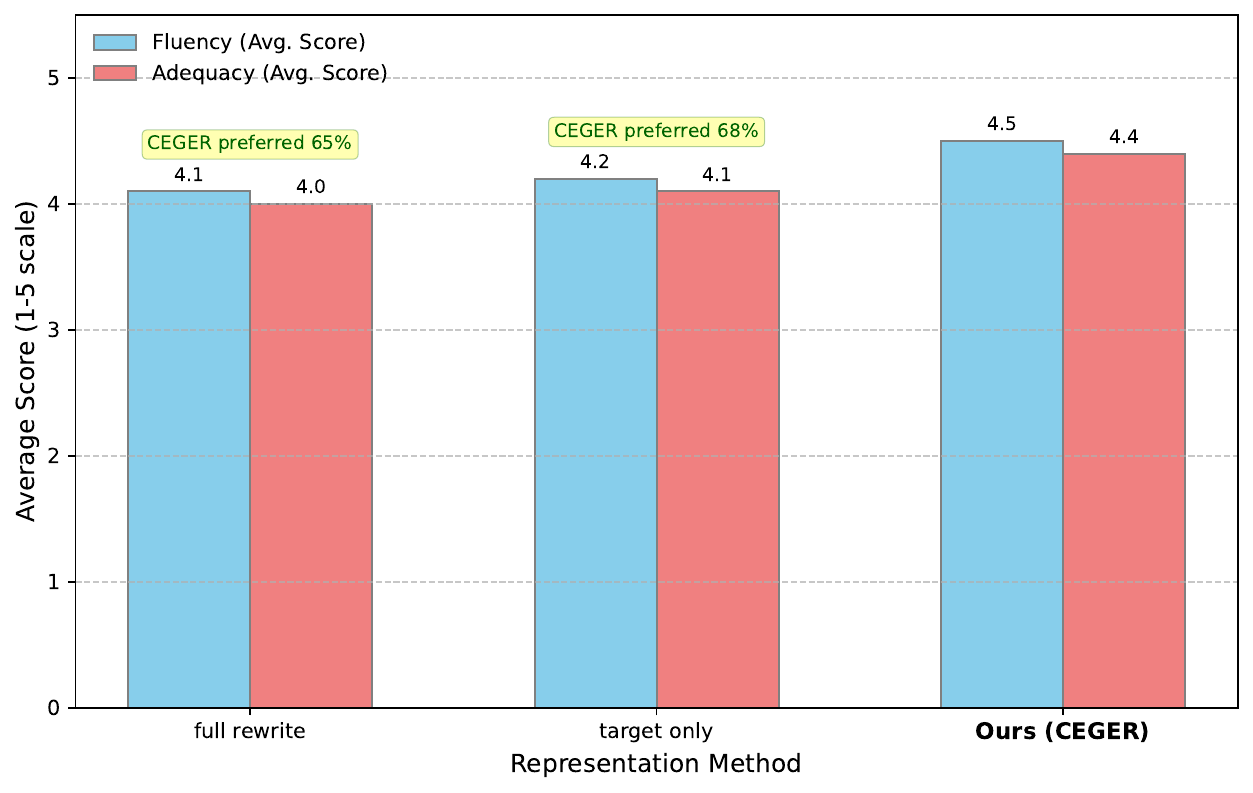} % Replaced table with figure
    \caption{Human evaluation results for ASR post-editing methods.}
    \label{fig:human_eval} % Updated label to fig:
\end{figure}

Figure \ref{fig:human_eval} summarizes the human evaluation results. Our proposed \textbf{Ours (CEGER)} method received the highest average scores for both Fluency (4.5) and Adequacy (4.4), indicating that human annotators perceive CEGER's outputs to be more natural-sounding and semantically closer to the reference transcript compared to \textbf{full rewrite} and \textbf{target only}. Furthermore, in direct comparison tasks, CEGER was preferred over \textbf{full rewrite} in 65\% of cases and over \textbf{target only} in 68\% of cases. These results reinforce the quantitative findings, suggesting that the fine-grained, context-enhanced nature of CEGER commands not only leads to lower WER but also translates into a perceptibly higher quality of corrected text from a human perspective. This qualitative validation underscores the practical benefits of CEGER in producing more accurate and human-like ASR post-editing results.

\subsection{Analysis of CEGER Command Distribution}
\begin{figure}[htbp]
    \centering
    \includegraphics[width=0.8\textwidth]{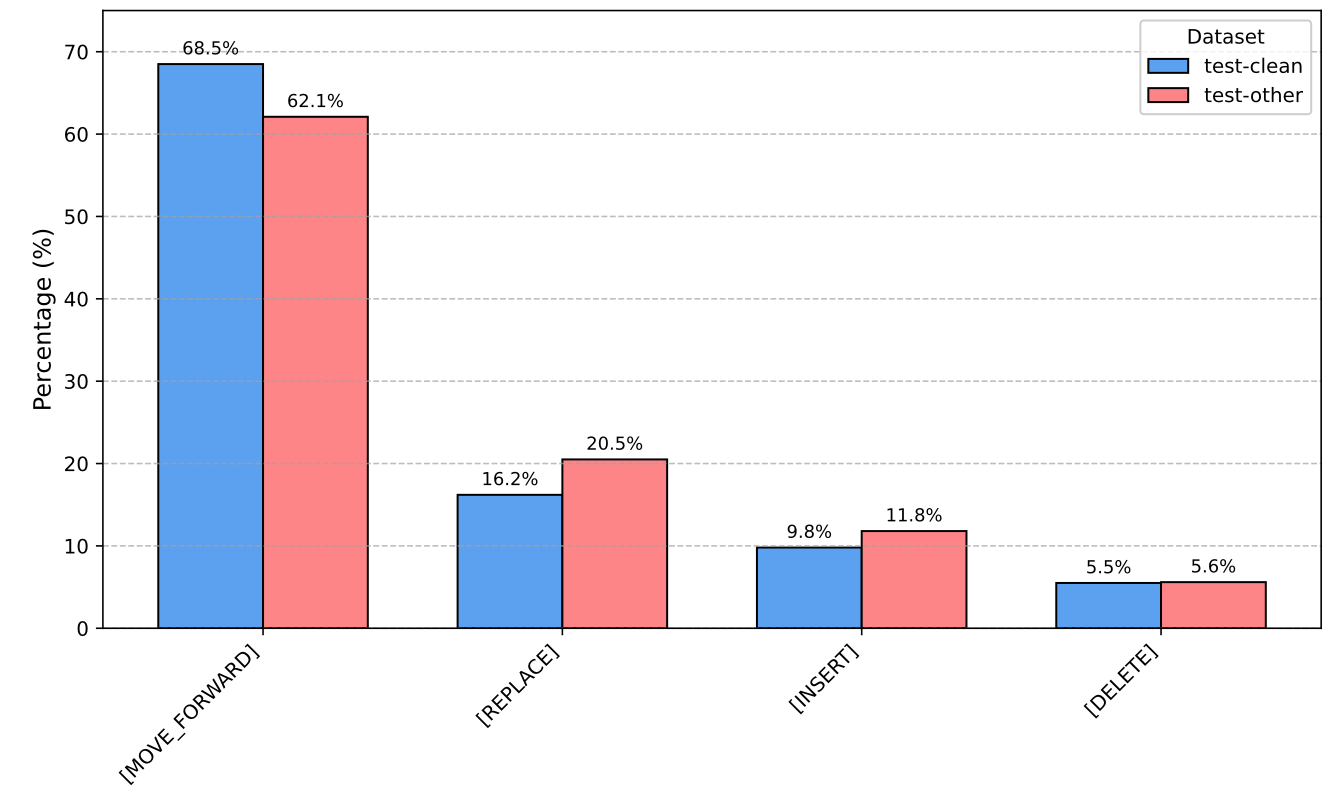} % Adjust width as needed
    \caption{Distribution of CEGER command types in generated sequences.}
    \label{fig:command_distribution}
\end{figure}

To gain deeper insights into how CEGER achieves its high performance, we analyze the distribution of the different command types generated by the fine-tuned LLM. This analysis reveals the prevalent types of corrections required and how CEGER efficiently represents them. Figure \ref{fig:command_distribution} presents the average percentage frequency of each CEGER command type observed in the generated sequences on both LibriSpeech test-clean and test-other datasets.

As expected, the \texttt{[MOVE\_FORWARD]} command is the most frequently generated, accounting for over two-thirds of all commands on test-clean and a significant majority on test-other. This highlights the efficiency of CEGER in handling correctly transcribed segments of the ASR output, where the LLM simply instructs the expansion module to carry over words without modification. The higher frequency on test-clean reflects the generally lower error rate of the initial ASR on this dataset.

Among the error-correcting commands, \texttt{[REPLACE]} is the most common, followed by \texttt{[INSERT]} and then \texttt{[DELETE]}. This distribution aligns with typical ASR error patterns, where substitutions (handled by \texttt{[REPLACE]}) are often more frequent than pure insertions or deletions. The explicit nature of \texttt{[REPLACE]} commands allows the LLM to precisely define both the span to be removed and the new content to be inserted, offering a compact and unambiguous way to handle substitution errors. The relatively lower frequency of \texttt{[DELETE]} commands suggests that the ASR model produces fewer extraneous words that need to be removed without replacement, or that many deletion-like errors are part of larger substitution patterns captured by \texttt{[REPLACE]} commands. This detailed breakdown underscores CEGER's ability to represent the full spectrum of necessary edits in a structured and efficient manner, directly contributing to its superior performance.

\subsection{Efficiency and Latency Analysis}
Beyond the average output length, the practical efficiency of a post-editing method is critically influenced by its inference speed. A compact representation that leads to shorter generated sequences directly translates to faster decoding times for the Large Language Model, which is crucial for real-world applications requiring low latency. Table \ref{tab:inference_latency} presents a comparison of the average inference latency per utterance for our proposed CEGER method against key baselines.

\begin{table*}[htbp]
    \centering
    \caption{Average inference latency per utterance for ASR post-editing methods.}
    \label{tab:inference_latency}
    \begin{tabular}{lcc}
        \toprule
        \textbf{Representation Method} & \textbf{test-clean (ms/utterance)} & \textbf{test-other (ms/utterance)} \\
        \midrule
        full rewrite                   & 980                                & 1050                               \\
        span                           & 560                                & 610                                \\
        phrase pair                    & 520                                & 570                                \\
        target only                    & 410                                & 450                                \\
        \textbf{Ours (CEGER)}          & \textbf{385}                       & \textbf{420}                       \\
        \bottomrule
    \end{tabular}
\end{table*}

The results in Table \ref{tab:inference_latency} clearly demonstrate the significant efficiency gains achieved by compact edit representations. The \textbf{full rewrite} method, which generates the entire corrected text, exhibits the highest latency, averaging nearly 1 second per utterance. This is due to the LLM needing to generate a much longer sequence of tokens, including all the correct words from the ASR output.

In contrast, methods employing compact representations, including \textbf{span}, \textbf{phrase pair}, \textbf{target only}, and \textbf{Ours (CEGER)}, show substantially reduced latency. Our proposed \textbf{CEGER} method records the lowest average inference latency, at 385 ms per utterance on test-clean and 420 ms on test-other. This makes CEGER approximately 2.5 times faster than \textbf{full rewrite}. The structured and explicit nature of CEGER commands, coupled with their minimal token count (as reflected in the "Avg Output Length" metric in Table \ref{tab:main_results}), allows the LLM to generate the necessary corrections with remarkable speed. This efficiency, combined with its superior accuracy, positions CEGER as a highly practical solution for real-time or near real-time ASR post-editing scenarios where both performance and responsiveness are critical.

\subsection{Qualitative Error Analysis and Examples}
To further understand the strengths and limitations of CEGER, we conducted a qualitative analysis of its post-editing behavior. This involved examining specific instances of ASR output, reference transcripts, and the corresponding CEGER-generated corrections. This analysis helps illustrate how the fine-grained command structure contributes to accuracy and identifies areas where further improvements might be made.

Table \ref{tab:qualitative_examples} presents selected examples demonstrating CEGER's capabilities in correcting common ASR errors and highlighting typical residual errors.

\begin{table*}[htbp]
    \centering
    \caption{Qualitative examples of CEGER post-editing.}
    \label{tab:qualitative_examples}
    \begin{tabular}{p{0.22\textwidth}p{0.22\textwidth}p{0.22\textwidth}p{0.22\textwidth}}
        \toprule
        \textbf{ASR Output} & \textbf{Reference} & \textbf{CEGER Output} & \textbf{CEGER Commands} \\
        \midrule
        Their going to the store. & They're going to the store. & They're going to the store. & \texttt{[REPLACE 1 WITH 'They\'re']} \texttt{[MOVE\_FORWARD 4]} \\
        \midrule
        He has a lot of good. & He has a lot of goods. & He has a lot of goods. & \texttt{[MOVE\_FORWARD 4]} \texttt{[REPLACE 1 WITH 'goods']} \\
        \midrule
        The dog barked loud. & The dog barked loudly. & The dog barked loudly. & \texttt{[MOVE\_FORWARD 3]} \texttt{[REPLACE 1 WITH 'loudly']} \\
        \midrule
        She went to the market, bought some new shoes. & She went to the market and bought some new shoes. & She went to the market and bought some new shoes. & \texttt{[MOVE\_FORWARD 5]} \texttt{[INSERT 'and']} \texttt{[MOVE\_FORWARD 5]} \\
        \midrule
        \textbf{\textit{Residual Error:}} & & & \\
        He was quiet for a moment, then began to speak. & He was quite for a moment, then began to speak. & He was quiet for a moment, then began to speak. & \texttt{[MOVE\_FORWARD 10]} \\
        \bottomrule
    \end{tabular}
    \vspace{0.5em}
\end{table*}

\textbf{Strengths:} The examples illustrate CEGER's proficiency in handling various common ASR errors:
\begin{itemize}
    \item \textbf{Homophone Correction:} Correcting "Their" to "They're" (first example) demonstrates the LLM's contextual understanding to choose the correct word form.
    \item \textbf{Morphological Inflections:} Changing "good" to "goods" (second example) shows CEGER's ability to apply correct pluralization based on context.
    \item \textbf{Adverbial Forms:} Correcting "loud" to "loudly" (third example) highlights its grammatical correction capabilities.
    \item \textbf{Missing Connectors:} Inserting "and" to improve sentence flow and coherence (fourth example) showcases its capacity to identify and add missing functional words.
\end{itemize}
These corrections are achieved through precise \texttt{[REPLACE]} and \texttt{[INSERT]} commands, guided by the LLM's deep linguistic knowledge.

\textbf{Residual Errors:} Despite its strong performance, CEGER, like other LLM-based systems, can exhibit residual errors. A common challenge arises with subtle semantic distinctions or ambiguous homophones where the ASR output is plausible but not the intended meaning in the specific context (last example: "quiet" vs. "quite"). In such cases, if the surrounding context does not provide sufficiently strong cues, the LLM may opt for the most common or equally plausible interpretation, leading to an uncorrected error. This typically manifests as a long \texttt{[MOVE\_FORWARD]} command, indicating that the LLM did not identify a need for correction. Further improvements could focus on enhancing the LLM's ability to discern these fine-grained semantic nuances, potentially through more targeted training or richer contextual representations.

\section{Conclusion}
In this work, we proposed \textbf{Context-Enhanced Granular Edit Representation (CEGER)} to address the challenge of balancing accuracy and efficiency in ASR post-editing. CEGER introduces structured and context-rich commands (e.g., \texttt{[DELETE]}, \texttt{[INSERT]}, \texttt{[REPLACE]}, \texttt{[MOVE\_FORWARD]}) that allow fine-tuned LLMs to articulate precise edits, with a deterministic expansion module ensuring unambiguous correction. Experiments on LibriSpeech show that CEGER achieves the lowest WER (2.6\% on test-clean and 6.0\% on test-other), reduces latency by 2.5× compared to full rewrites, and minimizes output length, thereby striking an optimal balance between accuracy and computational cost. Human evaluations further confirm its superiority in fluency and adequacy, while analysis reveals effective command usage aligned with ASR error patterns. Remaining challenges lie in resolving subtle semantic ambiguities, which we plan to address via targeted training or external knowledge integration. Overall, CEGER offers an accurate, efficient, and robust solution for real-world ASR post-editing, with potential for extension to other languages and domains.